\documentclass[runningheads]{llncs}

\usepackage[year=2026,ID=3925]{eccv}

\usepackage{algorithm} 
\usepackage{algpseudocode} 
\usepackage{amsmath}
\usepackage{soul}
\usepackage{booktabs}
\usepackage{overpic}
\usepackage{amssymb}
\usepackage{stfloats}
\usepackage{pifont}
\usepackage{multirow}
\usepackage{pifont}
\usepackage{colortbl}
\usepackage{subcaption}
\usepackage{comment}

\definecolor{mydarkred}{rgb}{0.6,0,0}
\definecolor{mydarkgreen}{rgb}{0,0.6,0}
\usepackage{tabularx}

\usepackage{eccvabbrv}

\usepackage{graphicx}
\usepackage{booktabs}

\usepackage[accsupp]{axessibility}  

\usepackage{hyperref}

\usepackage{orcidlink}

\makeatletter
\def\thanks#1{\protected@xdef\@thanks{\@thanks
        \protect\footnotetext{#1}}}
\makeatother

\begin{document}

\title{Open-Vocabulary 3D Object Detection \\ with Co-Distillation Discovery \\ and Dual Guidance Robust Training} 

\titlerunning{Co-3DGT}

\author{Shangbo Yuan\inst{1,\ddag}\thanks{$^{\ddag}$This work was carried out during Shangbo's visit to the IMPL Lab at SUTD. *Corresponding author.} \quad
Jie Xu\inst{2,*} \quad
Xiaofeng Zhu\inst{3} \quad
Na Zhao \inst{2}
}

\authorrunning{Yuan et al.}

\institute{
School of Computer Science and Engineering, \\ University of Electronic Science and Technology of China, Chengdu, China \and
Information Systems Technology and Design Pillar, \\ Singapore University of Technology and Design, Singapore \and 
School of Computer Science and Technology, Hainan University, Haikou, China
}

\maketitle

\begin{abstract}
Recently, open-vocabulary 3D object detection (3D-OVD) has gained increasing attention for its ability to detect unseen objects in 3D scenes. Existing approaches typically adopt a two-stage pipeline that first discovers novel objects using foundation models and then trains a 3D-OVD model based on these discovered objects. Although effective, this pipeline often suffers from inaccurate localization and mismatched classification during the discovery stage, which subsequently limits the performance of the model training stage. To address these limitations, we advocate for improving both the reliability of novel object discovery and the robustness of model training, and propose an innovative framework. Specifically, for reliable discovery, our co-distillation strategy distills high-quality novel objects by applying Hungarian matching over a comprehensive score that incorporates geometric consistency, structural objectness, and semantic certainty. To enhance robust model training, we further propose a dual-guidance learning scheme, incorporating a scene-awareness-guided uncertainty regularization for the regression head and an LLM-guided hierarchical alignment for the classification head, effectively mitigating the negative effects of imprecise 3D bounding boxes and semantic ambiguity. Extensive experiments on SUN RGB-D and ScanNetV2 demonstrate that our method achieves significant performance gains over state-of-the-art approaches. Code is available at \href{https://github.com/shangboyuan/Co-3DGT}{https://github.com/shangboyuan/Co-3DGT}.
\end{abstract}

\section{Introduction}\label{sec:Intro}
Open-vocabulary object detection (OVD)~\cite{zareian2021open,guopen,zhang2023simple} aims to localize and recognize novel objects beyond the training categories in unseen scenes by employing the unbounded vocabulary of vision-language foundation models~\cite{li2022blip,li2022grounded}.
Recently, OVD has achieved remarkable progress in 2D object detection tasks, largely driven by the foundation models (\eg, CLIP~\cite{radford2021learning}) pre-trained on massive 2D image-text datasets~\cite{deng2009imagenet,lin2014microsoft,everingham2010pascal,gupta2019lvis}.
However, extending these advances to open-vocabulary 3D object detection (3D-OVD)~\cite{etchegaray2024find,jiang2024open,deng20253dllava} remains challenging due to the intrinsic gap between 2D images and 3D point clouds~\cite{yuan2026graph}, as well as the limited availability of large-scale annotated 3D data, which hinders the training of 3D-specific foundation models.

\begin{figure}[t]
\centering
\includegraphics[page=1, width=1\linewidth]{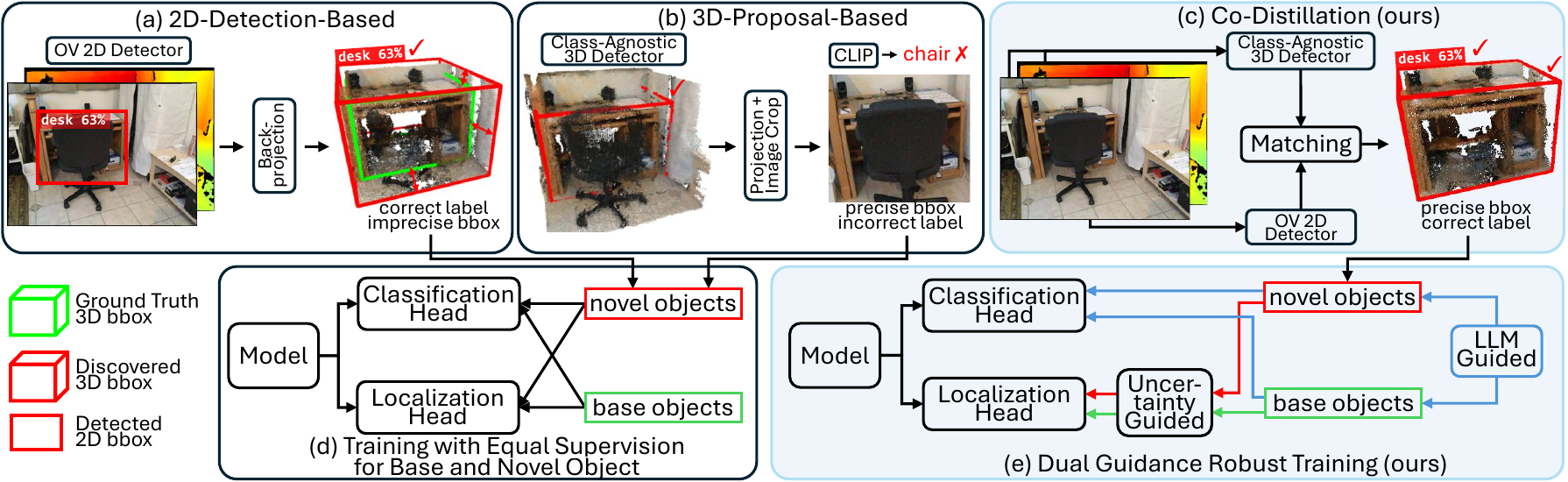}
\caption{\textbf{Comparison between our Co-3DGT and two existing 3D-OVD paradigms.}
In the \textit{Novel Object Discovery} stage:
(a) 2D-Detection-based strategy \cite{lu2023open,wang2024ov} back-projects 2D-OVD results into 3D space, often leading to imprecise 3D bounding boxes;
(b) 3D-Proposal-based methods \cite{cao2023coda,cao2025collaborative} project 3D proposals from class-agnostic 3D detectors onto 2D plane and crop images for open-vocabulary labeling, which may introduce semantic ambiguity;
(c) We propose Co-Distillation, jointly considering consider geometric constraints alongside 2D semantic and 3D localization inofrmation for more reliable novel objects. 
In the \textit{Training} stage:
(d) Previous methods typically employ equal supervision for both base and novel objects;
while
(e) We propose Dual Guidance Robust Training, integrating uncertainty regularization for the regression head and leveraging an LLM to achieve hierarchical semantic alignment.}
\label{fig:teaser}
\end{figure}

To achieve 3D-OVD, existing methods typically follow a two-stage pipeline: they first leverage 2D foundation models to discover novel objects in 3D scenes, and then train a 3D detector using both ground-truth base objects and discovered novel objects to transfer localization and classification knowledge to the 3D domain.
Depending on how novel object discovery stage is conducted, these methods can be categorized into two main types: \textit{2D-Detection-based} and \textit{3D-Proposal-based}.
As illustrated in Figure~\ref{fig:teaser}a, 2D-Detection-based methods~\cite{lu2023open,wang2024ov} perform open-vocabulary object detection on 2D images associated with a 3D scene. 
Specifically, a pre-trained OV 2D detector is utilized to identify target objects (\eg, a `desk' alongside its confidence score) within the 2D frame, 
leveraging global 2D contexts for visual recognition. 
Subsequently, these methods back-project the accurately detected 2D bounding boxes (bboxes) into the 3D space to form and localize the identified 3D objects.
In contrast, 3D-Proposal-based methods~\cite{cao2023coda,cao2025collaborative}, shown in Figure~\ref{fig:teaser}b, 
employ a class-agnostic 3D detector to generate 3D bboxes for all object proposals, which are projected onto the image plane to obtain corresponding 2D regions of interest. 
Vision-language models (\eg, CLIP~\cite{radford2021learning}) are then used to assign open-vocabulary class labels to the image crops of these detected 3D objects.
During the training stage, both the 3D bboxes and semantic labels of the discovered novel objects are employed to train a 3D-OVD model, 
where 3D bboxes are regressed for localization and semantic label features from CLIP are aligned for classification.

Despite the effectiveness of the two-stage pipeline, its performance is still limited by noise in the discovered novel objects, 
which negatively impacts the subsequent training stage that relies on them as supervision. 
While both 2D-Detection-based and 3D-Proposal-based methods inevitably introduce noise during the discovery stage, they suffer from two different major types of errors.
Specifically, the 2D-Detection-based strategy often yields inaccurate 3D bboxes because the noisy back-projection process encapsulates erroneous depth measurements and extraneous points from adjacent objects into the resulting 3D point set.
Thus, calculating a 3D bbox from such a noisy 3D point set leads to imprecise localization. 
For example, in Figure~\ref{fig:teaser}a, the back-projected bbox of the desk is largely imprecise because its underlying point set includes noisy and unassociated points.
On the other hand, 3D-Proposal-based approaches may assign incorrect semantic labels to 3D bboxes due to occlution, as shown in Figure~\ref{fig:teaser}b, 
the desk is partially occluded by the chair, causing the cropped image of the discovered proposal to be incorrectly classified as a chair by the CLIP model. 
Consequently, the discovery process directly bottlenecks the subsequent training stage, which relies on the bbox parameters and semantic labels from discovered novel objects for model optimization.

Moreover, the inherent errors stemming from both 2D and 3D detectors remain largely unavoidable in practical open-vocabulary scenarios. 
Specifically, 2D OVDs can also introduce semantic misclassifications (i.e., classification noise) \cite{tran2025simltd}, while class-agnostic 3D detectors inevitably yield imprecise spatial localization (i.e., bounding box noise) \cite{deng2025quantity}. 
However, during the subsequent training stage, existing methods typically treat the high-quality ground-truth annotations of base objects and the noisy pseudo-labels of discovered novel objects with \textit{equal supervision} as presented in Figure~\ref{fig:teaser}d.
Without accounting for the severe quality disparity between these two sources of annotation, localization accuracy of the model is impaired by the imprecise bboxes, 
and semantic errors from the 2D models are inadvertently propagated into the trained model.
This indiscriminate training strategy consequently degrades the feature representations for both base and novel categories, leading to suboptimal detection performance.

To address the above challenges, we propose open-vocabulary 3D object detection with \underline{Co-D}istillation \underline{D}iscovery and \underline{D}ual \underline{G}uidance \underline{T}raining (Co-3DGT), a unified framework that enhances both the \textit{reliability of novel 3D object discovery} and the \textit{robustness of model training}.
As illustrated in Figure~\ref{fig:teaser}c, our \emph{co-distillation discovery} module jointly leverages a 2D open-vocabulary detector and a class-agnostic 3D detector to discover novel objects from both spatial and semantic perspectives in parallel.
Specifically, we measure the spatial IoU scores between semantic 2D proposals and class-agnostic 3D proposals in a scene,
and then combine the 2D class confidence and 3D objectness scores to jointly solve a scene-level matching problem.
This objective expects to discover the spatial consistency among 3D boxes with semantic constraints, to co-distill reliable 3D novel objects.

To mitigate the impact of inevitable noise in discovered objects, we introduce the \textit{dual-guidance training strategy} as shown in Figure~\ref{fig:teaser}e,
which comprises (i) scene-awareness-guided uncertainty regularization and (ii) large language model (LLM)-guided hierarchical alignment. 
Concretely,
(i) we exploit the scene-aware uncertainty from base classes to regularize the uncertainties for novel classes, thus dynamically weighting the 3D bbox regression loss.
This prevents the model from becoming overconfident in noise 3D bboxes, thereby ensuring stable and robust localization performance.
Meanwhile,
(ii) we propose an LLM-guided hierarchical labeling to enrich label supervision instead of directly using given category labels.
It derives two levels of super-category labels from the original category names via an LLM to capture shared semantic and geometric properties among related classes.
Then, hierarchical semantic alignment aligns 3D object features with CLIP embeddings across multiple semantic levels, making the semantic classification less sensitive to noise labels.
This increases more discriminative supervision from super-categories to the original supervision, and thus helps the model to achieve robust classification.

In summary, our work makes four contributions as follows:
\begin{itemize}
\item We identify the performance bottleneck in the widely adopted two-stage 3D-OVD pipeline and propose a unified framework that jointly reduces noise in novel object discovery and enhances training robustness.
\item We propose a Co-Distillation Discovery strategy that enforces 3D-2D consistency via a joint optimization formulation that leverages spatial and semantic scores to selectively distill reliable novel 3D objects.
\item We introduce a Dual Guidance Robust Training scheme consisting of scene-awareness-guided uncertainty regularization for robust uncertainty-weighted regression, and LLM-guided hierarchical alignment for improved classification via label-augmented semantic feature alignment.
\item Extensive experiments on SUN RGB-D and ScanNetV2 demonstrate that our method achieves state-of-the-art performance, with accuracy improvements of 4.71\% and 9.82\% AP$_{25}$ in novel-class object detection, respectively.
\end{itemize}

\section{Related Work}

\noindent\textbf{3D Object Detection} 
focuses on identifying and localizing objects in 3D space from point clouds or RGB-D data~\cite{han2024dual,zhao2022static,wu2026ccf,zhao2020sess, sheng2025ct3d++}. 
VoteNet~\cite{qi2019deep} introduce an end-to-end point-based framework using Hough voting to generate anchor-free object proposals. Specifically, it employs PointNet++~\cite{qi2017pointnet++} to extract point cloud features, where seed points vote for object centers.
3DETR~\cite{misra2021end} adopts a Transformer architecture with self-attention for global context modeling, effectively eliminating the need for hierarchical aggregation and heuristic voting schemes. Instead, it uses learnable queries and a Hungarian matching loss for direct, end-to-end detection. 
More recently, Cubify-Anything~\cite{lazarow2025cubify} introduced the Cubify Transformer (CuTR) model alongside the Cubify-Anything 1M (CA-1M) dataset, further advancing indoor 3D object detection by leveraging Vision Transformer (ViT)-based~\cite{dosovitskiy2020vit} architectures.

\noindent \textbf{Open-Vocabulary Object Detection} seeks to generalize object detections beyond their seen training categories to recognize and localize previously unseen object classes by leveraging the powerful representational capabilities of vision-language pretraining models~\cite{alayrac2022flamingo,wang2023hierarchical,huang2024open,wang2025ov}.
The seminal work of CLIP~\cite{radford2021learning} establishes the foundational paradigm for this research field by learning semantically aligned vision-language representations through contrastive learning on web-scale image-text pairs. 
Based on this, Detic~\cite{zhou2022detecting} integrate zero-shot recognition into detectors like Faster R-CNN~\cite{ren2016faster} by replacing the fixed-vocabulary classification heads with CLIP-based classifiers that can dynamically accommodate arbitrary text object categories during inference without requiring additional fine-tuning.
Grounding DINO~\cite{liu2024grounding} achieves tighter vision-language integration by unifying detection and phrase grounding via language-guided query mechanisms, enabling the model to learn robust associations between textual descriptions and visual regions without requiring expensive box-level supervision annotations during the pretraining phase.

\noindent \textbf{Open-Vocabulary 3D Object Detection}
has emerged to overcome the scalability limitations of traditional closed-set detectors, which rely on exhaustive and costly 3D annotations~\cite{zhu2024open,wang2025uncertainty,zhu2026few,Xu_2026_CVPR,cao2025late}. 
OV-3DET~\cite{lu2023open} pioneers 3D-OVD without requiring 3D annotations by leveraging pretrained 2D detectors for semantic transfer. 
CoDA~\cite{cao2023coda} introduces a collaborative framework that integrates novel object discovery with cross-modal alignment, which jointly exploits 3D geometry to discover potential objects and 2D semantics for their identification, enhancing the model's ability to distinguish unknown from known instances.
Addressing the challenge of semantic ambiguity, INHA~\cite{jiao2024unlocking} advances this by seeding 3D novel object proposals from 2D OVD results and introducing a multi-level hierarchical alignment across instance, category, and scene contexts to reduce semantic inconsistency.
More recently, OV-Uni3DETR~\cite{wang2024ov} extends open-vocabulary detection to multimodal and outdoor environments, unifying representations across point clouds and RGB images within a transformer-based architecture, demonstrating robust performance and flexibility for various modality combinations.
\section{Method}\label{sec:method}
\noindent\textbf{Problem Definition.}
A 3D scene is represented by point cloud $\mathcal{P} =\{\mathbf{P} \in \mathbb{R}^{N \times 3} \}$ and corresponding RGB-D images $(I^\text{rgb} \in \mathbb{R}^{3 \times H \times W}, I^\text{depth} \in \mathbb{R}^{1 \times H \times W})$ with known camera intrinsics $\mathbf{K} \in \mathbb{R}^{3 \times 3}$, rotation and translation extrinsics $(\mathbf{R} \in \mathbb{R}^{3 \times 3}, \mathbf{t} \in \mathbb{R}^3)$.
Each 3D object in the scene is parameterized as $(\mathbf{b}, c)$, where $\mathbf{b}$ denotes the bbox defined as $\mathbf{b} = (x, y, z, l, w, h, \theta)$. Here, $(x,y,z)$ is the center coordinate, $(l,w,h)$ are the dimensions, and $\theta$ is the orientation around the vertical axis.
The category label is $c \in \mathcal{C}$, where $\mathcal{C}$ is the set of object categories.

Unlike closed-set 3D object detection, open-vocabulary 3D object detection is only provided with base objects $\mathcal{O}^{\text{base}} = \{\mathbf{b}_i, c_i \mid c_i \in \mathcal{C}^{\text{base}}\}$, while the goal is to detect and recognize objects from the combined category set $\mathcal{O} = \mathcal{O}^{\text{base}} \cup \mathcal{O}^{\text{novel}}$. To achieve this, a discovery stage is employed to identify potential novel objects ${\mathcal{O}}^{\text{novel}}= \{\mathbf{b}_j, c_j \mid c_j \in \mathcal{C}^{\text{novel}}\}$ within the scene. 
These discovered instances serve as additional pseudo-labels that expand the category coverage beyond $\mathcal{C}^{\text{base}}$.
In the training stage, both the annotated base objects $\mathcal{O}^{\text{base}}$ and the discovered novel objects ${\mathcal{O}}^{\text{novel}}$ are jointly utilized as supervision.
This 
enables the model to localize and recognize both base and novel objects.

\begin{figure*}[!t]
\centering
\includegraphics[page=1, width=1 \linewidth]{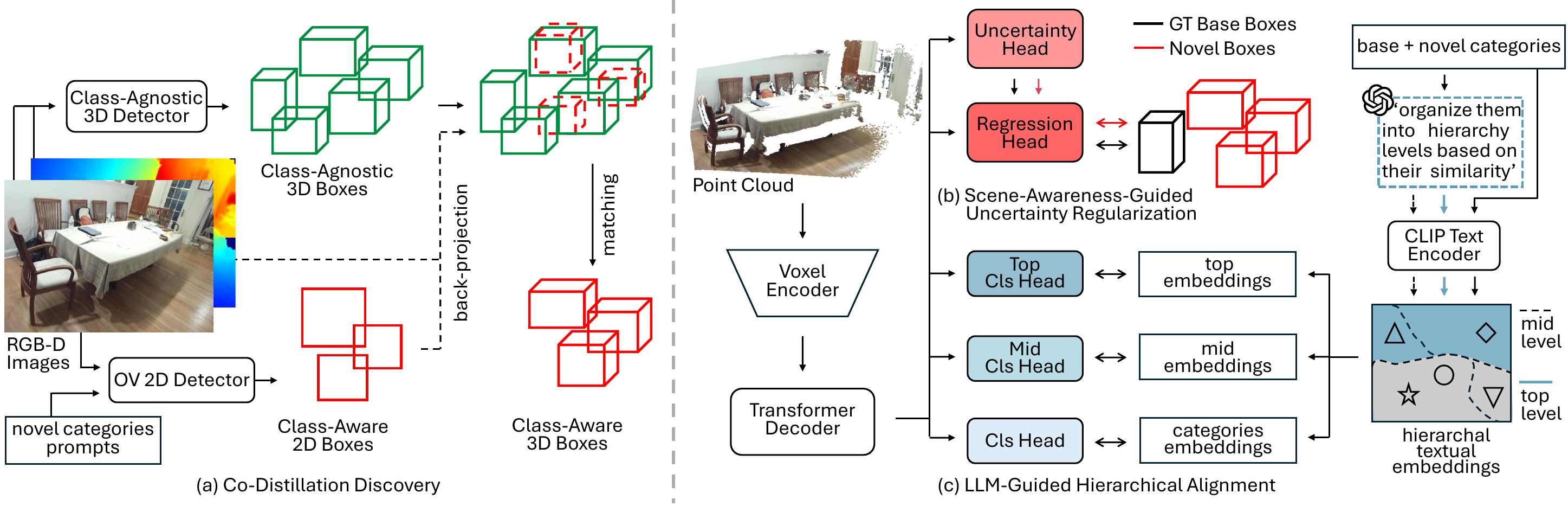}
\caption{\textbf{Framework Overview of our Co-3DGT.}
In \emph{Co-Distillation Discovery} stage: (a) a class-agnostic 3D detector generates 3D boxes, while a 2D open-vocabulary detector produces class-aware 2D boxes which are back-projected and to co-distill reliable class-aware 3D boxes.
In \emph{Dual Guidance Robust Training} stage: (b) \emph{Scene-Awareness-Guided Uncertainty Regularization} exploits the scene-aware uncertainty from base classes to regularize the uncertainties for novel classes, dynamically weighting the 3D bbox regression.
(c) \emph{LLM-Guided Hierarchical Alignment} leverage LLM to form super-categories that increase structural semantic supervision beyond original categories, promoting the model to achieve robust classification.}
\label{fig:framework}
\end{figure*}

\noindent\textbf{Framework Overview.}
As illustrated in Figure~\ref{fig:framework}, we propose Co-3DGT 
to enhance the reliability of novel object discovery and the robustness of model training.
Our framework consists of two stages.
In the novel object discovery stage, we propose a scene-level matching problem based co-distillation discovery strategy (Sec.~\ref{Co-Dis}) that can jointly distill high-quality novel objects.
In the model training stage, we propose a dual guidance robust training scheme, where a scene-awareness-guided uncertainty regularization (Sec.~\ref{Base-Guided}) and an LLM-guided hierarchical alignment (Sec.~\ref{LLM-Guided}) are designed to mitigate negative effects of the unavoidable imprecise 3D bboxes and semantic ambiguity, respectively.

\subsection{Co-Distillation Discovery}\label{Co-Dis}
We observe that the 2D-Detection-based discovery exhibits stronger performance in capturing semantically accurate object categories but suffers from producing noise 3D bboxes,
while the 3D-Proposal-based discovery is conducive to predicting precise 3D bboxes but is prone to yielding noise semantic labels.
Motivated by this, we propose a \emph{co-distillation discovery} approach to avoid the drawbacks of previous two pipelines, and to fully exploit the respective advantages of 3D localization and 2D semantic recognition.
That is, as shown in Figure~\ref{fig:framework}a, we first obtain semantically labeled 2D bounding boxes by leveraging the rich categorical knowledge of open-vocabulary 2D detectors, alongside class-agnostic 3D object proposals generated in parallel. 
We then introduce a bipartite matching problem, to jointly distill high-quality novel object annotations from both 2D semantic and 3D geometric perspectives.

Specifically, to lift 2D detections to 3D space, we follow the back-projection operation used in previous methods~\cite{lu2023open, wang2024ov}: 
$P_k = (K R_t)^{-1} \cdot (d_k \cdot \mathbf{x}_k)$.
The pixel set $\{\mathbf{x}_k\} = \{[u_k, v_k, 1]^\top\}$ within a 2D OVD result is back-projected to obtain the 3D point set $\{P_k\}$ using the camera intrinsic matrix $K$, camera extrinsics $R_t$, and the corresponding depth value $\{d_k\}$. 
The resulting 3D bbox is then defined as the minimal cuboid enclosing the 3D point set $\{P_k\}$.

Following this, let $\mathcal{O}^{\text{2D}} = \{ ({\mathbf{b}}_i^{\text{2D}}, c_i^{\text{2D}}, 
s_i^{\text{2D}}) \}_{i=1}^M$ denote the set of back-projected 3D objects derived from the 2D OVD results across all views. 
Here, ${\mathbf{b}}_i^{\text{2D}}$ represents the $i$-th 3D bbox lifted from the 2D OVD result, 
while $c_i^{\text{2D}}$ and $s_i^{\text{2D}}$ indicate its predicted semantic category label and associated confidence score, respectively.
Similarly, let $\mathcal{O}^{\text{3D}} = \{ ( \mathbf{b}_j^{\text{3D}}, s_j^{\text{3D}} ) \}_{j=1}^N$ represent the set of class-agnostic 3D proposals by the class-agnostic detector, 
where for the $j$-th proposal, $\mathbf{b}_j^{\text{3D}}$ denotes the 3D bbox and $s_j^{\text{3D}}$ is the foreground probability.

To achieve co-distillation discovery, we formulate the association between $M$ 2D bounding boxes and $N$ 3D proposals as a bipartite matching problem. 
To solve this matching problem via the Hungarian algorithm, we define the cost matrix $\mathbf{C} \in \mathbb{R}^{M \times N}$ as:
\begin{equation}
C_{ij} = -\mathrm{IoU}_{ij} - 0.1(s_j^{\text{3D}} + s_i^{\text{2D}}).
\end{equation}

Here, $\mathrm{IoU}_{ij}$ denotes the Intersection over Union (IoU) between the $i$-th 3D bbox lifted from the 2D OVD result and the $j$-th class-agnostic 3D proposal. 
To filter out geometrically unfeasible pairs, we establish a minimal IoU threshold $\tau = 0.1$. 
Specifically, if $\mathrm{IoU}_{ij} < \tau$, we prevent their assignment by setting the corresponding cost $C_{ij} = \infty$.

For candidates that satisfy this minimal IoU, their exact assignment is driven by minimizing the cost function.
The $\mathrm{IoU}_{ij}$ term serves as the primary geometric alignment metric, penalizing size mismatches and location misalignments, thereby encouraging tight spatial correspondence between cross-modal candidates. 
The 3D foreground probability $s_j^{\text{3D}}$ and the 2D confidence score $s_i^{\text{2D}}$ quantify the spatial objectness of the 3D proposal and the semantic reliability of the 2D prediction, respectively.
To preserve $\mathrm{IoU}_{ij}$ as the primary selection criterion, we apply a weight coefficient of 0.1 to both confidence scores.
Together, this composite cost drives the assignment of cross-modal pairs that are simultaneously geometrically consistent and semantically reliable, facilitating high-quality transfer of categorical knowledge from the 2D domain to the 3D representation.

By minimizing the total cost via the Hungarian algorithm, we obtain the optimal assignment set $\mathcal{H}$ of the matched pairs $(i, j)$. 
Consequently, our co-distillation strategy discovers and extracts more reliable novel objects to form the final distilled set as follows:
\begin{equation}
\mathcal{O}_{\text{Co-D}}^{\text{novel}} = \left\{ \left( \mathbf{b}_j^{\text{3D}}, c_i^{\text{2D}} \right) \mid (i, j) \in \mathcal{H} \right\}.
\end{equation}

Here, $\mathbf{b}_j^{\text{3D}}$ denotes the bbox parameters of the selected 3D proposal, and $c_i^{\text{2D}}$ represents the open-vocabulary semantic category transferred from its paired 2D detection. 
By pairing high-quality 3D geometry with holistically inferred 2D semantics, this formulation effectively constructs a more reliable pseudo-label set for novel classes, facilitating the subsequent learning process.

Given the base and discovered novel objects $\mathcal{O}^{} = \mathcal{O}^{\text{base}} \cup \mathcal{O}_{\text{Co-D}}^{\text{novel}}$,
we train a 3D detector to transfer the geometric localization and semantic classification knowledge to the 3D-OVD model.
Considering that both the class-agnostic 3D detector and the 2D open-vocabulary detector might introduce inevitable error, 
we propose the \textit{Dual Guidance Robust Training} scheme to achieve the robust 3D-OVD model with imperfect localization and classification supervision.
It consists of \textit{Scene-Awareness-Guided Uncertainty Regularization} and \textit{LLM-Guided Hierarchical Alignment} that will be introduced in following sections.

\subsection{Scene-Awareness-Guided Uncertainty Regularization}\label{Base-Guided} %

Inspired by the uncertainty estimation in deep learning~\cite{kendall2017uncertainties,seitzer2022pitfalls}, 
we employ the uncertainty-weighted regression loss to train the model as shown in Figure~\ref{fig:framework}b.
Specifically, the 3D bbox supervision for base objects (with manual annotation) and novel objects (from discovery stage) are derived from different processes, 
and they usually exhibit different noise levels.
To this end, we treat base and novel objects as two distinct supervision sources during training and design specialized regression loss functions for each.
For base objects, the bboxes $\mathbf{b}_i \in \mathcal{O}^{\text{base}}_{}$ are manually annotated, we adopt the following regression loss~\cite{seitzer2022pitfalls}:
\begin{align}
\label{eq:baseNLL}\mathcal{L}^\text{base}_\text{Reg}
&= \sum_{i} \left\lfloor \frac{\sigma_i}{\sqrt{2}} \right\rfloor^{\frac{1}{2}}\left(\frac{\sqrt{2}}{\sigma_i} , \bigl| \hat{\mathbf{b}}_i - \mathbf{b}_i \bigr| + \log \sigma^{\text{}}_i \right),
\end{align}
where $\hat{\mathbf{b}}_i $ is the $i$-th prediction of the 3D bbox of the model for the object $\mathbf{b}_i \in \mathcal{O}^{\text{base}}$, 
and $\sigma_i$ is the estimated uncertainty for the $i$-th object obtained by neural networks~\cite{kendall2017uncertainties}.
$\lfloor \cdot \rfloor$ denotes the stop-gradient operation.

In this formulation, the regression loss is driven by three key components. 
We first employ two complementary weighting terms, 
where the uncertainty weighting term ${\sqrt{2}}/{\sigma_i}$ enables the model to focus on reliable bboxes with low $\sigma_i$, 
and the adaptive re-weighting term $\left\lfloor {\sigma_i}/{\sqrt{2}} \right\rfloor^{\!\frac{1}{2}}$ prevents the model from ignoring difficult samples. 
Next, $\log \sigma_i$ acts as a regularization term, which explicitly penalizes the network for predicting an arbitrarily large uncertainty $\sigma_i$ to reduce loss.
Finally, the interplay among these three components effectively maintains balanced supervision across varying uncertainty levels for base boxes.

For novel objects, the bboxes $\mathbf{b}_j \in \mathcal{O}^{\text{novel}}_{\text{Co-D}}$ are generated by the class-agnostic 3D detector, and consequently contain additional localization errors.
The uncertainty term $\log \sigma_i$ in Eq.~\eqref{eq:baseNLL} is learned primarily from precise manual annotations of base objects.
As a result, the learned uncertainty estimates may not adequately capture the elevated noise level of novel bboxes.
To address this issue, we improve Eq.~\eqref{eq:baseNLL} and propose the scene-awareness-guided uncertainty regularization as follows:
\begin{equation}
\label{novelNLL}\mathcal{L}^\text{novel}_\text{Reg}= \sum_j \Biggl(\frac{\sqrt{2}}{\sigma_j}\bigl| \hat{\mathbf{b}}_j - \mathbf{b}_j \bigr|\bigl| \log \sigma_j - \mathbb{E}\left[ \lfloor \log \sigma_i \rfloor \right] \bigr|\Biggr),
\end{equation}
where $\mathbb{E} \left[ \lfloor \log \sigma_i \rfloor \right]$ represents the scene-aware inherent uncertainty predicted by the model, derived from the expectation of the stop-gradient log-uncertainty of base bboxes. 
To handle scenes without base objects, we maintain a moving average $\mathbb{E}'$ term updated using the expectations from base-containing scenes via a standard momentum update (decay rate of $0.9$) during training.

\textcolor{black}{Here, the term $|\log \sigma_j - \mathbb{E}[\lfloor \log \sigma_i \rfloor]|$ serves as a soft regularization that guides the uncertainty of novel objects toward the scene-aware uncertainty. 
The motivation of our regression loss is that we hope the bbox uncertainty of novel objects remains at the same level as that of base objects 
$\mathbb{E}\left[ \lfloor \log \sigma_{\text{base}} \rfloor \right]$, which can be supported by the classical theory of maximum mean discrepancy (MMD).}
This prevents both underestimation (when $\log \sigma_j \leq \mathbb{E} \left[ \lfloor \log \sigma_i \rfloor \right]$) and unbounded overestimation (when $\log \sigma_j > \mathbb{E} \left[ \lfloor \log \sigma_i \rfloor \right]$).Notably, we omit the adaptive re-weighting term $\left\lfloor {\sigma}/{\sqrt{2}} \right\rfloor^{\!\frac{1}{2}}$ from Eq.~\eqref{eq:baseNLL} since applying this term to novel bboxes with higher noise levels could amplify erroneous signals. 

Overall, the 3D bbox regression loss with our scene-awareness-guided uncertainty regularization is:
\begin{align}
\mathcal{L}_{\text{Reg}}
&= \mathcal{L}^\text{base}_\text{Reg}+\mathcal{L}^\text{novel}_\text{Reg}.
\end{align}

This scene-awareness-guided uncertainty regularization leverages scene-aware uncertainty estimates from base objects as an anchor to robustly guide the uncertainty prediction for novel objects within the same scene.

\subsection{LLM-Guided Hierarchical Alignment}\label{LLM-Guided}

To achieve the robustness against inaccurate semantic information form the given category labels,
we first propose an LLM-guided hierarchical labeling to enrich label supervision, and then conduct the hierarchical semantic alignment to achieve the reliable classification as shown in Figure~\ref{fig:framework}c.

\noindent\textbf{LLM-Guided Hierarchical Labeling.}
To mitigate the possible noise supervision from the given category labels, we employ an LLM
to infer functional, geometric, and conceptual relationships among base and novel categories, to obtain the label-augmented supervision.
Specifically, we define three hierarchies for each category label, \ie, top, mid, bottom.
The bottom level contains the original fine-grained category labels $\mathcal{C}^{\text{bot}} = \mathcal{C}^{\text{base}} \cup \mathcal{C}^{\text{novel}}$, while the top and mid levels $\mathcal{C}^{\text{mid}},~\mathcal{C}^{\text{top}}$ are super-categories inferred by the LLM.
For example, hierarchical labels from $\mathcal{C}^{\text{bot}}$ is constructed as:
\begin{equation}
\left\{
\begin{aligned}
& \mathcal{C}^{\text{top}}: \text{`furniture', `appliance', `accessory', ...} \\
& \mathcal{C}^{\text{mid}}: \text{`seating', `storage', `domestic', `tools', ...} \\
& \mathcal{C}^{\text{bot}}: \text{`chair', `table', `pillow', `door', `fridge', ...} 
\end{aligned}
\right.
\end{equation}

In our method, the mid level corresponds to functionally and geometrically similar super-categories, and the top level represents abstract ones.
The top and mid levels are less sensitive to semantic ambiguity from visually similar objects, which provides structural information for subsequent semantic alignment.

\noindent\textbf{Hierarchical Semantic Alignment.}
For the $j$-th semantic label, we use CLIP to encode its corresponding hierarchical labels $c^{\text{top}}_j$, $c^{\text{mid}}_j$, $c^{\text{bot}}_j$ to obtain the textual embeddings $\mathbf{h}_j^{(1)}$, $\mathbf{h}_j^{(2)}$, $\mathbf{h}_j^{(3)}$, respectively.
These embeddings are subsequently aligned with the 3D-OVD model which also produce hierarchical semantic features $\mathbf{f}_i^{(1)}$, $\mathbf{f}_i^{(2)}$, $\mathbf{f}_i^{(3)}$.

To obtain the predicted probability that the $i$-th sample belongs to the $j$-th category at level $l$, we first compute the cosine similarity score and then map it to the interval (0,1) using a sigmoid function:
\begin{equation}
p_{ij}^{(l)} = \text{sigmoid}\left( \frac{\langle \mathbf{f}_i^{(l)},\mathbf{h}_j^{(l)} \rangle}{\Vert \mathbf{f}_i^{(l)} \Vert \Vert \mathbf{h}_j^{(l)} \Vert} \right),
\end{equation}
and then adopt binary cross-entropy (BCE) loss:
\begin{equation}
\mathcal{L}_\text{bce}^{(l)}=  - \frac{1}{N} \sum_{j}^{|\mathcal{C}^{(l)}|}  \sum_{i}^{N}  \Big[ y_{ij}^{(l)} \cdot \log(p_{ij}^{(l)}) + (1 - y_{ij}^{(l)}) \cdot \log(1 - p_{ij}^{(l)}) \Big]
\end{equation}
where $y_{ij}^{(l)}$ denotes the indicator variable, e.g., $y_{ij}^{(l)} = 1 $ if the $i$-th object belongs to the $j$-th category at the $l$-th level, otherwise $y_{ij}^{(l)} = 0 $.

The overall hierarchical alignment loss is formulated as:

\begin{equation}
\mathcal{L}_\text{Align} = \mathcal{L}_\text{bce}^{(1)} +  \mathcal{L}_\text{bce}^{(2)} + \mathcal{L}_\text{bce}^{(3)}.
\end{equation}

In the hierarchical semantic alignment, the bottom level preserves fine-grained category distinctions and supervises the learning of detailed object features, while the mid and top levels mitigate the impact of noise labels by improving robustness against visually similar or ambiguous objects.

\section{Experiments}

\subsection{Experimental Setup}
\noindent\textbf{Datasets.}
We evaluate our proposed method on two challenging large-scale indoor 3D object detection datasets: SUN RGB-D~\cite{song2015sun} and ScanNetV2~\cite{dai2017scannet}. The SUN RGB-D dataset contains about 10,335 RGB-D scenes across diverse indoor environments. Following common practice, we use 5,285 training samples covering 46 object categories, where the 10 most frequent categories are treated as base classes and the remaining 36 as novel classes.
The ScanNetV2 dataset~\cite{dai2017scannet} consists of more than 1500 indoor scenes with 200 object categories. From the 1,201 training scenes in ScanNetV2, we select the 10 most frequent classes as base classes and the next 50 as novel classes. In addition, following OV-3DET~\cite{lu2023open}, we adopt a setting where no ground-truth annotations are provided, and evaluate the top 20 most frequent unseen classes on ScanNetV2.

\noindent\textbf{Baselines.}
We compare our method with a series of 3D-OVD methods that take point clouds as input. 
These baselines include Det-PointCLIP~\cite{yao2022detclip}, Det-PointCLIPv2~\cite{yao2023detclipv2}, Det-CLIP$^2$~\cite{zeng2023clip2}, OV-3DET~\cite{lu2023open}, CoDA~\cite{cao2023coda}, INHA~\cite{jiao2024unlocking}, CoDAv2~\cite{cao2025collaborative}, and OV-Uni3DETR~\cite{wang2024ov}.

\noindent\textbf{Evaluation Metrics.}
We adopt the mean Average Precision (mAP) and mean Average Recall (mAR) at an IoU threshold of 0.25 for evaluation. 
We report the mAP for unseen classes, seen classes, and all classes, denoted as 
$\text{AP}^{\text{novel}}_{25}$, $\text{AP}^{\text{base}}_{25}$, and $\text{AP}^{\text{mean}}_{25}$, respectively. 
Similarly, the corresponding mAR are denoted as
$\text{AR}^{\text{novel}}_{25}$, $\text{AR}^{\text{base}}_{25}$, and $\text{AR}^{\text{mean}}_{25}$.
In addition, under the same setting with no ground-truth annotations following OV-3DET, we adopt the $\text{AP}^{\text{}}_{25}$ as the metric for the top 20 most frequent unseen classes.

\noindent\textbf{Implementation Details.}
We adopt the backbone of Uni3DETR~\cite{wang2023uni3detr}, including its voxel encoder and transformer decoder, as our 3D detection model. 
In line with common approaches~\cite{yao2022detclip,zeng2023clip2,wang2024ov,cao2023coda,jiao2024unlocking,lu2023open},
we utilize CLIP~\cite{radford2021learning} to transform category labels into textual features. 
Following previous work~\cite{wang2024ov,jiao2024unlocking,lu2023open}, we leverage Detic as the 2D open-vocabulary detector. 
Additionally, we employ CuTR~\cite{lazarow2025cubify} as the 3D class-agnostic detector and ChatGPT-5~\cite{singh2025openai} to infer category relationships and generate hierarchical labels. 
In terms of computational cost, our full pipeline requires approximately 47 minutes and 169 minutes for novel object discovery on SUN RGB-D and ScanNetV2, respectively, using a single RTX 3090 GPU.
The LLM-guided hierarchical alignment introduces negligible time cost, requiring only a single API call for the entire dataset (approximately 20 $\sim$ 30 seconds) to generate the two-level hierarchical labels.
Although the discovery stage is moderately slower than OV-Uni3DETR (25 minutes and 103 minutes with single RTX 3090 GPU on SUN RGB-D and ScanNetV2, respectively), our approach substantially reduces the overall training time. 
Using four RTX 3090 GPUs, the training stage takes 490 and 309 minutes on SUN RGB-D and ScanNetV2, respectively (compared to 716 and 621 minutes for OV-Uni3DETR), since our model operates only on point cloud inputs, in contrast to OV-Uni3DETR which integrates both point cloud and image modalities.

\subsection{Main Results}
As shown in Table \ref{tab:open_vocab_3d}, our method achieves the highest overall performance on both SUN RGB-D and ScanNetV2 benchmarks. Notably, our model significantly improves detection of novel categories, boosting $\text{AP}^{\text{novel}}_{25}$ to 14.37\% on SUN RGB-D and 21.91\% on ScanNetV2, which corresponds to considerable gains of +4.71\% and +9.82\% over the previous state-of-the-art approach (OV-Uni3DETR). In comparison, the improvements on base categories are more modest, with increases of +1.56\% on SUN RGB-D and +2.36\% on ScanNetV2. Overall, our model achieves a $\text{AP}^{\text{mean}}_{25}$ of 22.63\% on SUN RGB-D and 23.75\% on ScanNetV2, surpassing OV-Uni3DETR by +4.57\% and +8.60\%, respectively. These results demonstrate that our approach effectively enhances robustness to unseen categories, while maintaining strong performance on base categories.

Additionally, in the same annotation-free setting following OV-3DET~\cite{lu2023open}, no ground-truth annotated base objects are available. As a result, our scene-awareness-guided uncertainty regularization becomes inapplicable. In this scenario, our model is trained using a simplified variant that retains only the LLM-guided hierarchical alignment with the novel objects discovered by Co-Distillation. Our method still achieves 36.15\% $\textup{AP}^{\text{mean}}_{25}$ as shown in Table~\ref{tab:per_class}, improving upon OV-Uni3DETR by +10.82\%. This improvement confirms the reliability of our Co-Distillation discovery and the robustness of the LLM-guided hierarchical alignment.

\begin{figure*}[t]
    \centering
    \begin{subfigure}{0.24\linewidth}
        \centering
        \includegraphics[page=1,width=\linewidth]{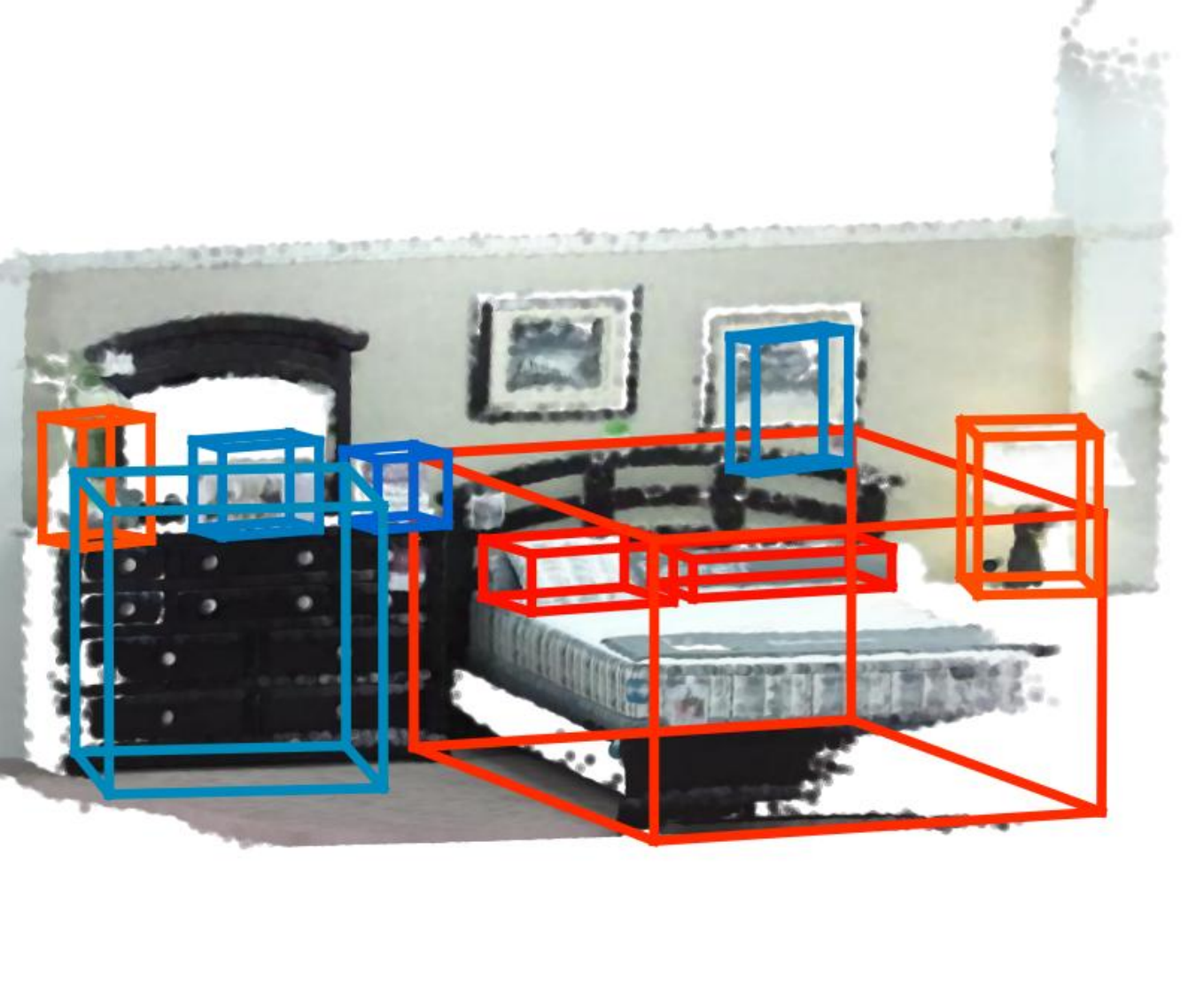}
        \caption{}
        \label{fig:vis_a}
    \end{subfigure}
    \begin{subfigure}{0.24\linewidth}
        \centering
        \includegraphics[page=2,width=\linewidth]{fig/vis.pdf}
        \caption{}
        \label{fig:vis_b}
    \end{subfigure}
    \begin{subfigure}{0.24\linewidth}
        \centering
        \includegraphics[page=3,width=\linewidth]{fig/vis.pdf}
        \caption{}
        \label{fig:vis_c}
    \end{subfigure}
    \begin{subfigure}{0.24\linewidth}
        \centering
        \includegraphics[page=4,width=\linewidth]{fig/vis.pdf}
        \caption{}
        \label{fig:vis_d}
    \end{subfigure}
    \begin{subfigure}{0.24\linewidth}
        \centering
        \includegraphics[page=5,width=\linewidth]{fig/vis.pdf}
        \caption{}
        \label{fig:vis_e}
    \end{subfigure}
    \begin{subfigure}{0.24\linewidth}
        \centering
        \includegraphics[page=6,width=\linewidth]{fig/vis.pdf}
        \caption{}
        \label{fig:vis_f}
    \end{subfigure}
    \begin{subfigure}{0.24\linewidth}
        \centering
        \includegraphics[page=7,width=\linewidth]{fig/vis.pdf}
        \caption{}
        \label{fig:vis_g}
    \end{subfigure}
    \begin{subfigure}{0.24\linewidth}
        \centering
        \includegraphics[page=8,width=\linewidth]{fig/vis.pdf}
        \caption{}
        \label{fig:vis_h}
    \end{subfigure}
\caption{Qualitative examples on SUN RGB-D (the first row) and ScanNetV2 (the
second row) to visualize the 3D-OVD results by our method. The red and orange colored bboxes correspond to different base-class objects, while blue and green colored bboxes represent different novel-class objects.}
    \label{fig:vis_result}
\end{figure*}

Moreover, Figure~\ref{fig:vis_result} presents qualitative examples of 3D-OVD results by our method in various indoor scenes on SUN RGB-D and ScanNetV2. Each subfigure illustrates detected 3D objects within reconstructed indoor scenes, indicating our method's effectiveness to localize and recognize both base and novel objects.

\begin{table*}[t]
\caption{Comparison of $\textup{AP}_{25}$ (\%) and $\textup{AR}_{25}$ (\%) for open-vocabulary 3D object detection on SUN RGB-D and ScanNetV2 datasets.}
\label{tab:open_vocab_3d}
\centering
\small
\renewcommand{\arraystretch}{1.0}
\setlength{\tabcolsep}{1pt}
\resizebox{\linewidth}{!}{%
\begin{tabular}{l|cccccc|cccccc}
\toprule[2pt]
\multirow{2}{*}{\textbf{Method}} & 
\multicolumn{6}{c|}{\textbf{SUN RGB-D}} & 
\multicolumn{6}{c}{\textbf{ScanNetV2}} \\ 
& $\textup{AP}^{\text{novel}}_{25}$ & $\textup{AP}^{\text{base}}_{25}$ & $\textup{AP}^{\text{mean}}_{25}$ 
& $\textup{AR}^{\text{novel}}_{25}$ & $\textup{AR}^{\text{base}}_{25}$ & $\textup{AR}^{\text{mean}}_{25}$ 
& $\textup{AP}^{\text{novel}}_{25}$ & $\textup{AP}^{\text{base}}_{25}$ & $\textup{AP}^{\text{mean}}_{25}$ 
& $\textup{AR}^{\text{novel}}_{25}$ & $\textup{AR}^{\text{base}}_{25}$ & $\textup{AR}^{\text{mean}}_{25}$ \\
\midrule
Det-PointCLIP~\cite{yao2022detclip} & 0.09 & 5.04 & 1.17 & 21.98 & 65.03 & 31.33 & 0.13 & 2.38 & 0.50 & 33.38 & 54.88 & 36.96 \\
Det-PointCLIPv2~\cite{yao2023detclipv2} & 0.12 & 4.82 & 1.14 & 21.33 & 63.74 & 30.55 & 0.13 & 1.75 & 0.40 & 32.60 & 54.52 & 36.25 \\
Det-CLIP$^2$~\cite{zeng2023clip2} & 0.88 & 22.74 & 5.63 & 22.21 & 65.04 & 31.52 & 0.14 & 1.76 & 0.40 & 34.26 & 56.22 & 37.92 \\
3D-CLIP~\cite{radford2021learning} & 3.61 & 30.56 & 9.47 & 21.47 & 63.74 & 30.66 & 3.74 & 14.14 & 5.47 & 32.15 & 54.15 & 35.81 \\
CoDA~\cite{cao2023coda} & 6.71 & 38.72 & 13.66 & 33.66 & 66.42 & 40.78 & 6.54 & 21.57 & 9.04 & 43.36 & 61.00 & 46.30 \\
INHA~\cite{jiao2024unlocking} & 8.91 & 42.17 & 16.18 & 51.34 & 78.65 & 57.23 & 7.79 & 25.10 & 10.68 & 55.10 & 71.60 & 57.85 \\
CoDAv2~\cite{cao2025collaborative} & 9.17 & 42.04 & 16.31 & 43.16 & 71.64 & 49.35 & 9.12 & 23.35 & 11.49 & 64.00 & 72.16 & 65.36 \\
OV-Uni3DETR~\cite{wang2024ov} & 9.66 & 48.29 & 18.06 & -- & -- & -- & 12.09 & 30.47 & 15.15 & -- & -- & -- \\
\rowcolor{gray!10}
\textbf{Co-3DGT (ours)} & 
\textbf{14.37} & \textbf{49.85} & \textbf{22.63} & 
\textbf{57.70} & \textbf{79.36} & \textbf{63.08} & 
\textbf{21.91} & \textbf{32.83} & \textbf{23.75} & 
\textbf{66.30} & \textbf{74.21} & \textbf{69.33} \\
\bottomrule[2pt]
\end{tabular}
}
\end{table*}

\begin{table*}[t]
\caption{Detailed comparison of per-category $\textup{AP}_{25}$ (\%) for open-vocabulary 3D object detection on ScanNetV2 dataset across 20 unseen categories.}
\label{tab:per_class}
\centering
\small
\renewcommand{\arraystretch}{1.0}
\setlength{\tabcolsep}{1pt}
\resizebox{\linewidth}{!}{%
\begin{tabular}{l|c|ccccccccccc}
\toprule[2pt]
\textbf{Method} 
& ~Mean~
& ~~toilet~ 
& ~~~bed~~~ 
& ~~chair~~ 
& ~~sofa~~~ 
& ~dresser~ 
& ~~table~~ 
& ~cabinet~ 
& bookshelf 
& ~pillow~~ 
& ~~sink~~~ 
\\
\midrule
OV-3DET~\cite{lu2023open}     & 
18.02 & 57.29 & 42.26 & 27.06 & 31.50 & 8.21 & 14.17 & 2.98 & 5.56 & 23.00 & 31.60 \\
CoDA~\cite{cao2023coda}       & 
19.32 & 68.09 & 44.04 & 28.72 & 44.57 & 3.41 & 20.23 & 5.32 & 0.03 & 27.95 & 45.26 \\
CoDAv2~\cite{cao2025collaborative}       & 
22.72 & 77.24 & 43.96 & 15.05 & 53.27 & 11.37 & 13.96 & 1.42 & 0.11 & 34.42 & 44.38 \\
OV-Uni3DETR~\cite{wang2024ov} & 
25.33 & 86.05 & 50.49 & 28.11 & 31.51 & 18.22 & 24.03 & 6.58 & 12.17 & 29.62 & 54.63 \\
\rowcolor{gray!10} \textbf{Co-3DGT (ours)} & 
\textbf{36.15} & 92.75 & 70.42 & 81.27 & 60.64 & 12.50 & 39.15 & 15.92 & 12.56 & 37.58 & 60.59 \\
\midrule[2pt]
\textbf{Method}  &  
& bathtub 
& refrigerator 
& desk 
& nightstand 
& counter 
& door 
& curtain 
& box 
& lamp 
& bag 
\\
\midrule
OV-3DET~\cite{lu2023open}     &
& 56.28 & 10.99 & 19.72 & 0.77 & 0.31 & 9.59 & 10.53 & 3.78 & 2.11 & 2.71 \\
CoDA~\cite{cao2023coda}       &
& 50.51 & 6.55 & 12.42 & 15.15 & 0.68 & 7.95 & 0.01 & 2.94 & 0.51 & 2.02 \\
CoDAv2~\cite{cao2025collaborative}       &
& 55.60 & 24.41 & 20.67 & 20.72 & 0.28 & 13.54 & 0.92 & 4.16 & 4.37 & 9.20 \\
OV-Uni3DETR~\cite{wang2024ov} &
& 63.73 & 14.41 & 30.47 & 2.94 & 1.00 & 19.02 & 19.90 & 12.70 & 5.58 & 13.46 \\
\rowcolor{gray!10} \textbf{Co-3DGT (ours)} &
& 69.52 & 34.57 & 24.23 & 13.15 & 7.01 & 15.99 & 16.86 & 15.99 & 38.61 & 4.35 \\
\bottomrule[2pt]
\end{tabular}
}
\end{table*}

\begin{table*}[t]
\caption{Ablation study of the Co-Distillation module on SUN RGB-D and ScanNetV2. 
Models are evaluated using $\textup{AP}^{\text{novel,disc}}_{25}$ (\%) and $\textup{AR}^{\text{novel,disc}}_{25}$ (\%) to measure pseudo-label quality on the training samples, 
and $\textup{AP}^{\text{novel,det}}_{25}$ (\%) and $\textup{AR}^{\text{novel,det}}_{25}$ (\%) to assess final detection performance.}
\centering
\resizebox{\linewidth}{!}{
\begin{tabular}{l|cc|cc|cc|cc}
\toprule[2pt]
\multirow{3}{*}{\textbf{Method}} & \multicolumn{4}{c|}{\textbf{Novel Object Discovery}} & \multicolumn{4}{c}{\textbf{Training Result}} \\
\cmidrule(){2-5} \cmidrule(){6-9}
& \multicolumn{2}{c|}{\textbf{SUN RGB-D}} & \multicolumn{2}{c|}{\textbf{ScanNetV2}} & \multicolumn{2}{c|}{\textbf{SUN RGB-D}} & \multicolumn{2}{c}{\textbf{ScanNetV2}} \\
& $\textup{AP}^{\text{novel,disc}}_{25}$ 
& $\textup{AR}^{\text{novel,disc}}_{25}$ 
& $\textup{AP}^{\text{novel,disc}}_{25}$ 
& $\textup{AR}^{\text{novel,disc}}_{25}$ 
& $\textup{AP}^{\text{novel,det}}_{25}$ 
& $\textup{AR}^{\text{novel,det}}_{25}$ 
& $\textup{AP}^{\text{novel,det}}_{25}$ 
& $\textup{AR}^{\text{novel,det}}_{25}$ \\
\midrule
2D-Detection-based & 6.21 & 16.10 & 3.93 & 23.44 & 9.59 & 52.99 & 11.92 & 57.85 \\
3D-Proposal-based   & 8.49 & 12.02 & 4.57 & 20.28 & 9.88 & 52.15 & 12.71 & 58.38 \\
Co-distil(scratch)   
& 10.45(\textcolor{mydarkred}{+1.96}) & 19.72(\textcolor{mydarkred}{+7.70}) 
& 9.35(\textcolor{mydarkred}{+4.76}) & 28.46(\textcolor{mydarkred}{+8.18}) 
& 11.76(\textcolor{mydarkred}{+1.88}) & 54.26(\textcolor{mydarkred}{+2.11}) 
& 16.68(\textcolor{mydarkred}{+3.97}) & 63.70(\textcolor{mydarkred}{+5.32}) \\
Co-distil(CuTR) 
& 10.97(\textcolor{mydarkred}{+2.48}) & 24.33(\textcolor{mydarkred}{+12.31}) 
& 10.46(\textcolor{mydarkred}{+5.89}) & 32.29(\textcolor{mydarkred}{+12.01}) 
& 12.69(\textcolor{mydarkred}{+2.81}) & 55.41(\textcolor{mydarkred}{+3.26}) 
& 18.74(\textcolor{mydarkred}{+6.03}) & 64.19(\textcolor{mydarkred}{+5.81}) \\
\bottomrule[2pt]
\end{tabular}
}
\label{tab:discovery}
\end{table*}

\begin{table*}[t]
\centering
\small
\renewcommand{\arraystretch}{1.0}
\setlength{\tabcolsep}{6pt}
\parbox{0.51\linewidth}{
\centering
\caption{Ablation study for dual guidance robust model training. $\textup{AP}^{\text{mean}}_{25}$ (\%) on SUN RGB-D and ScanNetV2 are reported.}
\label{tab:component}
\resizebox{\linewidth}{!}{%
\begin{tabular}{l|p{2.3cm}<{\centering}|p{2.1cm}<{\centering}}
\toprule[2pt]
\textbf{Variants} & \textbf{SUN RGB-D} & \textbf{ScanNetV2} \\
\midrule
Training w/o dual guidance                       & 20.62  & 20.15  \\
SGUR \ding{51}                  & 22.17 (\textcolor{mydarkred}{+1.55}) & 23.12 (\textcolor{mydarkred}{+2.97}) \\
LGHA \ding{51}                  & 21.35 (\textcolor{mydarkred}{+0.73}) & 22.73 (\textcolor{mydarkred}{+2.58}) \\
SGUR \ding{51} + LGHA \ding{51} & 22.63 (\textcolor{mydarkred}{+2.01}) & 23.75 (\textcolor{mydarkred}{+3.60}) \\
\bottomrule[2pt]
\end{tabular}
}
}
\hfill
\parbox{0.45\linewidth}{
\centering
\caption{Ablation study for LLM selection in LGHA.
$\textup{AP}^{\text{mean}}_{25}$ (\%) on SUN RGB-D and ScanNetV2 are reported.}
\label{tab:LLM}
\resizebox{\linewidth}{!}{%
\begin{tabular}{l|p{2.15cm}<{\centering}|p{1.6cm}<{\centering}}
\toprule[2pt]
\textbf{Models} & \textbf{SUN RGB-D} & \textbf{ScanNetV2} \\
\midrule
ChatGPT-5               & 22.63 & 23.75 \\
Gemini-3-Flash-Thinking  & 21.89 & 23.84 \\
Qwen3.5-397B-A17B        & 22.71 & 23.34 \\
Llama-3.3-70B-Instruct   & 21.42 & 23.51 \\
\bottomrule[2pt]
\end{tabular}
}
}
\end{table*}

\subsection{Ablation Study}
In Tables~\ref{tab:discovery}, ~\ref{tab:component} and \ref{tab:LLM}, we conduct ablation studies to verify the effectiveness of each component in our framework.

\noindent\textbf{Effectiveness of Co-Distillation Discovery.}
We first evaluate the performance of different discovery pipelines on the training set, 
focusing on the instance-level precision and recall for novel categories (left side of Table~\ref{tab:discovery}).  
Compared to the 2D-Detection-based and 3D-Proposal-based pipelines, our Co-Distillation discovery pipeline yields notable improvements. 
Specifically, compared to the 3D-Proposal-based pipeline, Co-distil(scratch) improves precision by +0.97\% and recall by +3.86\% on SUN RGB-D, alongside gains of +4.97\% and +8.18\% on ScanNetV2.
Incorporating the pretrained class-agnostic 3D detector, CuTR, further enhances discovery capability. 
Co-distil(CuTR) improves precision by +2.99\% and recall by +10.72\% on SUN RGB-D, over the 3D-Proposal-based baseline, and achieves +5.89\% and +12.01\% respective gains on ScanNetV2.

Subsequently, we analyze how the discovery results from Co-Distillation with CuTR translate to the final detection performance for novel classes (right side of Table~\ref{tab:discovery}), 
where our proposed dual-guidance training strategies are not applied. Better quality in novel object discovery directly contributes to higher final training results. 
Relying solely on the baseline training method without our robust training methods, Co-distil(CuTR) achieves 12.69\% $\textup{AP}^{\text{novel}}_{25}$ and 55.41\% $\textup{AR}^{\text{novel}}_{25}$ on SUN RGB-D, as well as 18.74\% $\textup{AP}^{\text{novel}}_{25}$ and 64.19\% $\textup{AR}^{\text{novel}}_{25}$ on ScanNetV2. 
These consistent improvements across both the training set and the final detection results verify that our Co-Distillation effectively mines high-quality novel objects to boost model performance.

\noindent\textbf{Effectiveness of Dual Guidance Robust Training.}
Based on the results of Co-Distil(CuTR) discovery, we further ablate our proposed dual guidance robust training strategies in Table~\ref{tab:component}, including scene-awareness-guided uncertainty regularization (SGUR) and LLM-guided hierarchical alignment (LGHA). 
Here, we focus on the performance improvements across all classes ($\textup{AP}^{\text{mean}}_{25}$). 
On SUN RGB-D, applying SGUR alone improves the performance by +1.55\% (achieving 22.17\% $\textup{AP}^{\text{mean}}_{25}$), while LGHA alone provides a +0.73\% gain (reaching 21.35\% $\textup{AP}^{\text{mean}}_{25}$). 
When combined, they bring a total improvement of +2.01\%, yielding a final result of 22.63\% $\textup{AP}^{\text{mean}}_{25}$. We observe a similar trend on ScanNetV2, where SGUR and LGHA contribute gains of +2.97\% and +2.58\%, respectively. 
Their combination achieves 23.75\% $\textup{AP}^{\text{mean}}_{25}$, surpassing the baseline by +3.60\%. 
These results demonstrate the complementary roles of SGUR and LGHA in mitigating noise and enhancing overall training robustness.

\noindent\textbf{Influence of Different LLMs.}
In Table~\ref{tab:LLM}, we investigate the impact of utilizing different LLMs to generate the hierarchical class labels required for LGHA. Our results indicate that changing the underlying LLM yields similar performances of $\textup{AP}^{\text{mean}}_{25}$ with minimal fluctuations. This demonstrates that our LGHA strategy is robust to the choice of the LLM, as the alignment relies primarily on the rich structural and semantic hierarchy of the generated class labels rather than the specific capabilities or phrasing of a particular LLM.

\section{Conclusion}
In this work, we propose an innovative framework for 3D-OVD, focusing on reliable novel object discovery and robust model training. For reliable discovery, we introduce a Hungarian matching-based co-distillation strategy to generate high-quality novel objects by integrating 2D semantic and 3D geometric information. For robust model training, we introduce a dual guidance training method including scene-awareness-guided uncertainty regularization for robust localization and LLM-guided hierarchical alignment for mitigating semantic ambiguity in classification. Experiments show significant performance gains over state-of-the-art methods. In future work, we plan to extend the proposed framework toward real-time 3D-OVD by incorporating temporal fusion to enable efficient perception for robotic navigation in indoor environments. 

\section*{Acknowledgments}
This research was supported in part by the National Key Research \& Development Program of China under Grant 2022YFA1004100,
and in part by the Ministry of Education, Singapore, under its MOE Academic Research Fund Tier 2 (MOE-T2EP20124-0013).

\bibliographystyle{splncs04}
\bibliography{main}

\end{document}